\documentclass[runningheads]{llncs}
\usepackage{geometry}
\usepackage{graphicx}
\usepackage{booktabs} 
\usepackage{graphicx}
\usepackage{subfigure}
\usepackage{lipsum}
\usepackage{xcolor}
\usepackage{hyperref}
\newcommand{\ev}[1]{{\color{red}#1}}

\usepackage{lipsum}
\usepackage{changepage}
\usepackage{array}
\usepackage{booktabs}
\usepackage{multirow}

\usepackage[linesnumbered,vlined,ruled,commentsnumbered]{algorithm2e}
\usepackage{algpseudocode}
\SetKwRepeat{Do}{do}{while}

\SetAlFnt{\small}
\SetAlCapFnt{\small}
\SetAlCapNameFnt{\small}
\IncMargin{-\parindent}
\SetKw{Break}{break}

\definecolor{lightyellow}{RGB}{255,255,224}
\definecolor{lightgreen}{RGB}{215,238,145}
\definecolor{lightpurple}{RGB}{230,230,250}
\definecolor{lightgrey}{RGB}{211,211,211}
\definecolor{lightblue}{RGB}{173,216,230}

\usepackage{tcolorbox}
\tcbuselibrary{listings,breakable}
\usepackage{changepage} 

\usepackage{listings} \usepackage{xcolor}

 \usepackage{booktabs} 

\usepackage{tcolorbox}
\usepackage{listings}

\definecolor{bg}{gray}{0.95}
\DeclareTCBListing{mintedbox}{O{}m!O{}}{%
  breakable=true,
  listing engine=minted,
  listing only,
  minted language=#2,
  minted style=default,
  minted options={%
    linenos,
    gobble=0,
    breaklines=true,
    breakafter=,,
    fontsize=\small,
    numbersep=8pt,
    #1},
  boxsep=0pt,
  left skip=0pt,
  right skip=0pt,
  left=25pt,
  right=0pt,
  top=3pt,
  bottom=3pt,
  arc=5pt,
  leftrule=0pt,
  rightrule=0pt,
  bottomrule=2pt,
  toprule=2pt,
  colback=bg,
  colframe=orange!70,
  enhanced,
  overlay={%
    \begin{tcbclipinterior}
    \fill[orange!20!white] (frame.south west) rectangle ([xshift=20pt]frame.north west);
    \end{tcbclipinterior}},
  #3}

\usepackage{listings}
\usepackage{tcolorbox}
\usepackage{xcolor}

\definecolor{bg}{rgb}{0.95,0.95,0.95}

\usepackage{listings}
\usepackage{xcolor} 

\begin{document}

\title{Iterative Resolution of Prompt Ambiguities Using a Progressive Cutting-Search Approach}

\titlerunning{Resolution of Prompt Ambiguities Using a Pruning Approach}

\author{Fabrizio Marozzo\inst{1} \orcidID{0000-0001-7887-1314}} 

\authorrunning{F. Marozzo}

\institute{University of Calabria, Italy\\
\email{fmarozzo@dimes.unical.it}
}
\maketitle            
\begin{abstract}
Generative AI systems have revolutionized human interaction by enabling natural language-based coding and problem solving. However, the inherent ambiguity of natural language often leads to imprecise instructions, forcing users to iteratively test, correct, and resubmit their prompts. We propose an iterative approach that systematically narrows down these ambiguities through a structured series of clarification questions and alternative solution proposals—illustrated with input/output examples as well. Once every uncertainty is resolved, a final, precise solution is generated. Evaluated on a diverse dataset spanning coding, data analysis, and creative writing, our method demonstrates superior accuracy, competitive resolution times, and higher user satisfaction compared to conventional one-shot solutions, which typically require multiple manual iterations to achieve a correct output.
\end{abstract}

\keywords{Ambiguity Resolution \and Interactive Prompting \and Generative AI \and Prompt Engineering \and Natural Language Processing}

\section{Introduction}
\label{sec:intro}

    The rapid adoption of generative AI systems is fundamentally reshaping how humans interact with technology. Widely used models from leading IT companies, such as OpenAI's GPT-4 and other cutting-edge solutions developed by companies like Google and DeepMind, have demonstrated advanced reasoning capabilities and the ability to articulate their thought processes during interactions \cite{akhtar2024unveiling,wang2023interactive}. This enables users to engage with technology through natural language, seamlessly solving a wide array of tasks—including creative writing and data analysis—in an intuitive and conversational manner \cite{devlin2019bert}. By not only delivering answers but also providing insights into their reasoning, these models empower users of all skill levels to work more efficiently and make better-informed decisions.
    
    In the field of coding, these systems have become invaluable. They not only assist programmers in writing and understanding code but also help diagnose and resolve errors \cite{chen2021evaluating}. This shift is leading to a new paradigm: strong declarative programming through natural language prompts. Traditionally, declarative programming has involved using formal languages to specify what a program should accomplish without detailing the underlying control flow. Languages like SQL for database queries, HTML/CSS for web layouts, and various high-level configuration languages exemplify this approach—users define the desired outcome, and the system figures out the steps to achieve it \cite{nikolaeva2024survey}. However, these languages require users to learn specific syntaxes and semantics, which can be a barrier for many.
    
    Building on this shift, AI-driven systems are now making coding even more accessible by allowing users to articulate their intent in everyday language. This natural language approach democratizes programming by reducing the steep learning curve associated with formal languages \cite{raffel2019exploring,ouyang2022training}. The AI interprets the user's intent and translates these prompts into executable code, enabling both programmers and non-programmers to harness the power of automation more intuitively and efficiently, thereby accelerating development cycles and making technology more accessible \cite{bommasani2021foundation,radford2019language}. However, this convenience introduces a significant challenge: natural language is inherently imprecise and fraught with ambiguities. Unlike a formal algorithm—defined by a clear, unambiguous list of instructions—natural language can be interpreted in multiple ways \cite{kaplan2020scaling}. As anyone with programming experience knows, ambiguous requirements often necessitate an iterative process of refinement: a provisional solution is provided, the user tests it, and subsequent adjustments are made until the final, correct implementation is achieved \cite{huang2019ai}.
    
    To address this challenge, we propose an iterative approach that systematically resolves ambiguities through a structured series of clarification questions. The process begins by analyzing the user’s input to identify potential ambiguities. For each issue detected, the system engages the user in a dialogue, presenting alternative solutions—often illustrated with input/output examples—to clarify the intended meaning. As ambiguities are resolved, the system dynamically adjusts the remaining uncertainties. If resolving one ambiguity clarifies or eliminates others, they are automatically removed from the process. This iterative refinement continues until all ambiguities have been addressed. At this stage, the system generates a final, precise solution that accurately reflects the user’s intent. To further validate the solution, the system provides representative examples, including edge cases, to illustrate the main behavior of the code. These examples allow users to assess how the final solution performs under various conditions, ensuring that the results align with their expectations.
    
    To evaluate the effectiveness of our iterative ambiguity resolution process, we conducted a study using a diverse dataset of natural language prompts across three domains: coding, data analysis, and creative writing. We assessed the system’s ability to accurately identify and resolve ambiguities, comparing it to conventional large language models that generate one-shot solutions without interactive clarification. Our evaluation focused on three key aspects: accuracy, efficiency, and user experience. The results demonstrate that our iterative approach significantly improves the precision of generated outputs by systematically refining ambiguous prompts, ensuring that the final solution better aligns with user intent. Additionally, despite requiring an initial clarification phase, our method maintains competitive resolution times by reducing the number of failed attempts and revisions typically needed in one-shot prompting. User feedback further highlights the advantages of guided disambiguation, reporting higher satisfaction due to improved clarity and reduced trial-and-error effort.

This paper is structured as follows: Section~\ref{sec:related} reviews related work on ambiguity resolution and prompt engineering. Section~\ref{sec:approach} details our proposed iterative solution based on a progressive cutting-search approach. Section~\ref{sec:experiments} presents our experimental results, including representative examples and evaluations. Finally, Section~\ref{sec:conclusion} concludes the paper by summarizing our contributions and outlining directions for future research.

\section{Related Work}
\label{sec:related}

Recent advances in large-scale language models (LLMs) have revolutionized multiple domains by demonstrating remarkable capabilities in natural language processing tasks, including coding \cite{liu2024your}, data analysis \cite{tian2024spreadsheetllm}, and creative writing \cite{fan2018hierarchical}. Systems such as GPT-4, Claude, Gemini, and other state-of-the-art large language models leverage massive datasets and complex architectures to generate contextually relevant, coherent, and often insightful outputs across diverse applications. These models not only deliver high-quality responses but also provide explanations of their reasoning processes, thereby enhancing interactive experiences between humans and machines \cite{brown2020language,chowdhery2022palm}.

In the realm of software development, the code generation capabilities of LLMs have had a profound impact. Tools such as OpenAI Codex (powering GitHub Copilot), Amazon CodeWhisperer, Google AlphaCode, DeepMind AlphaDev, and Tabnine assist developers in writing, understanding, and debugging code. These systems leverage advanced AI models to generate multi-line functions, optimize existing code, and suggest improvements, streamlining the development process and reducing the entry barrier for new programmers \cite{nguyen2023generative}.

For example, Chen et al. \cite{chen2021evaluating} provide empirical evidence that LLMs are capable of producing coherent and functional code segments tailored to specific requirements, thereby reducing the manual effort typically needed for debugging and refinement. Li et al. \cite{li2024approach} propose a method that leverages prompt engineering to rapidly generate source code, demonstrating that dynamically optimized prompts can significantly enhance the consistency and quality of generated code. Choi et al. \cite{choi2023consistency} further explore how prompt-based approaches can improve code comprehension and ensure that the intended functionality is preserved, addressing common issues like ambiguity in code behavior. Additionally, Wang et al. \cite{wang2023prompttuning} show that prompt tuning can outperform traditional fine-tuning methods, particularly in low-resource scenarios, by aligning model outputs more closely with developer expectations. Collectively, these studies underscore the transformative role of LLMs and prompt engineering in modern software development, paving the way for more efficient, reliable, and accessible coding practices.

Ambiguity in natural language poses significant challenges to machine understanding, particularly in tasks such as code generation and data analysis where precision is critical. Recent studies have explored iterative disambiguation techniques that engage users in clarification dialogues, thereby systematically resolving ambiguities. These approaches emphasize the importance of interactive systems that adapt to user feedback and progressively narrow down multiple interpretations until a precise and unambiguous instruction is achieved \cite{yadav2021comprehensive}. 
For example, He et al. \cite{he2025enhancing} introduced a human–machine co-adaptation framework that leverages multi-turn dialogues and targeted clarifying questions to iteratively refine ambiguous prompts, ensuring that the intended meaning is progressively clarified. Similarly, Aina and Linzen \cite{aina2021language} investigated how language models maintain and adjust syntactic uncertainty by sampling multiple completions, effectively quantifying the extent of ambiguity in a prompt. Moreover, best practices in prompt engineering, as outlined by Sabit Ekin \cite{ekin2023prompt}, advocate for iterative prompt refinement and the explicit specification of constraints to mitigate vagueness and improve output quality. Collectively, these studies underscore that integrating interactive, iterative disambiguation processes is essential to enhance the precision and contextual alignment of model responses across a variety of applications.

Unlike conventional approaches that rely on trial-and-error refinements, our method introduces a structured iterative clarification framework that systematically resolves ambiguities before generating a final response. Rather than producing an initial output that may require multiple manual corrections, our approach engages users in a guided resolution process, dynamically eliminating invalid interpretations and refining intent through targeted clarification steps. This ensures that the final output is both precise and aligned with user expectations while minimizing the need for post-hoc adjustments and reducing overall interaction time.

\section{Proposed approach}
\label{sec:approach}

\begin{figure}[htb!]
	\centering
	\includegraphics[width=1\linewidth]{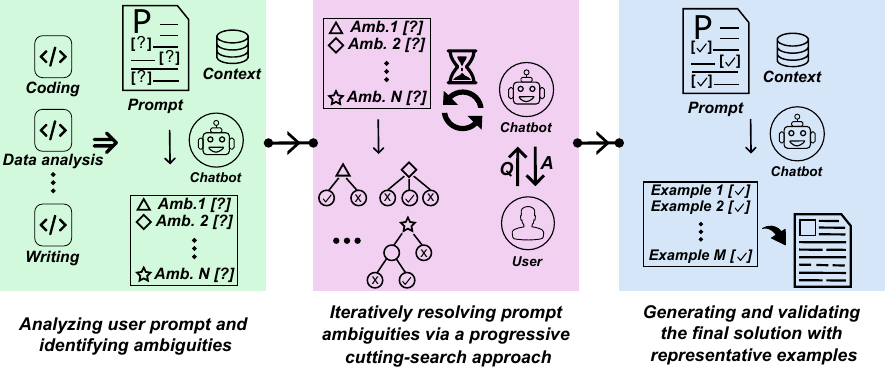}
	\caption{Execution flow of the proposed approach.
 \ev{}}
	\label{fig:proposed_approach}
\end{figure} 

The proposed approach is aimed at transforming ambiguous natural language prompts into precise, executable solutions by systematically identifying and resolving uncertainties. It comprises three distinct phases: (i) \textit{detecting ambiguities in the prompt}; (ii) \textit{iteratively resolving these ambiguities through a decision-tree dialogue with the user}; and (iii) \textit{generating and validating the final solution with representative examples, including edge cases}. In the following, we provide a detailed description of the main steps of our approach, whose execution flow is depicted in Figure~\ref{fig:proposed_approach}.

In the initial phase—\textit{identifying ambiguities in the prompt}—the system conducts a thorough analysis of the user's natural language input to pinpoint potential sources of misunderstanding. By leveraging advanced natural language processing techniques, including a chatbot such as GPT-4o via API, it detects ambiguous terms and phrases that could lead to multiple interpretations. Additionally, the system considers contextual information provided by the user, such as domain-specific constraints, to refine the ambiguity detection process. This allows the system to differentiate between inherently vague expressions and terms that may be clear within a specific context. By incorporating both linguistic analysis and contextual cues, this step lays the foundation for a more precise and targeted resolution process.

The second phase focuses on \textit{iteratively resolving these ambiguities via a progressive cutting-search approach}. Here, the system engages the user in a structured dialogue using a chatbot model to present alternative interpretations for each detected ambiguity, along with illustrative input/output examples. The dialogue follows a progressive cutting-search strategy, in which each user response eliminates invalid interpretations, dynamically refining the search space until a precise and unambiguous meaning is reached. This iterative narrowing process ensures that every uncertainty is addressed while minimizing unnecessary interactions. By guiding the system through successive clarifications, users efficiently arrive at a well-defined prompt that accurately reflects their intent.

In the final phase—\textit{generating and validating the final solution with representative examples}—the system compiles the resolved interpretations into a definitive solution, ensuring that all ambiguities have been systematically addressed. To further validate the accuracy and robustness of the output, the system, again leveraging a chatbot generates representative examples, including edge cases and borderline scenarios, that illustrate the main behavior of the final solution. These examples allow users to assess whether the solution meets their specific requirements, performs consistently under different conditions, and aligns with their expectations. If necessary, users can provide additional feedback to refine the solution before finalization, ensuring a high degree of reliability and adaptability across various contexts.

\section{Experimental Results}
\label{sec:experiments}

In this section, we present the experimental results of our iterative ambiguity resolution system. Our evaluation assesses the system's effectiveness across diverse domains, comparing its performance to standard one-shot methods. We report both quantitative metrics—including ambiguity detection accuracy, resolution time, and user satisfaction—and qualitative analyses of representative outputs. The results demonstrate that our guided, interactive approach not only improves the precision of generated outputs but also significantly reduces the time and effort required from users.

Further details on these findings are provided in Section~\ref{sec:examples}, which presents specific examples of ambiguity resolution across different domains, and Section~\ref{sec:eval}, which offers an overview of the system's performance based on aggregated results and key evaluation metrics.

\subsection{Case Studies and Input Examples}
\label{sec:examples}

As a first case study, we consider a generic transactional database commonly used in e-commerce systems. This database consists of two primary tables: \texttt{orders} and \texttt{customers}. The \texttt{orders} table contains fields such as \texttt{order\_id}, \texttt{order\_date}, \texttt{customer\_id}, \texttt{total\_amount}, and \texttt{order\_status}, capturing essential details about each transaction. The \texttt{customers} table includes fields like \texttt{customer\_id}, \texttt{name}, \texttt{email}, \texttt{registration\_date} and \texttt{total\_spent}, providing information about customer identity, activity, and purchasing behavior.

Below is an example of a user-provided prompt that will be examined for ambiguities when converting natural language into SQL. This prompt serves as the starting point of our process.

\begin{tcolorbox}[breakable=true,boxsep=0pt,left=2mm,right=2mm,top=2mm,bottom=2mm, sharp corners, colback=lightyellow, colframe=black, boxrule=0.5pt, fontupper=\small] \texttt{Prompt}: Translate the following request into SQL:
\textit{Find all orders placed last month by customers with high spending habits.} \end{tcolorbox}\vspace{0.2cm}

The next box lists the ambiguities detected in the prompt. These ambiguities indicate the areas that require further clarification.

\begin{tcolorbox}[breakable=true,boxsep=0pt,left=2mm,right=2mm,top=2mm,bottom=2mm, sharp corners, colback=lightgreen, colframe=black, boxrule=0.5pt, fontupper=\small] \texttt{Ambiguities}:

\textbf{A1}: Time Frame Interpretation (``\textit{last month}");

\textbf{A2}: Definition of ``\textit{high spending habits}". \end{tcolorbox}\vspace{0.2cm}

Next, we present a sequence of clarifying questions designed to resolve these ambiguities. Each step in this dialogue helps narrow down the possible interpretations.

\begin{tcolorbox}[breakable=true,boxsep=0pt,left=2mm,right=2mm,top=2mm,bottom=2mm, sharp corners, colback=lightpurple, colframe=black, boxrule=0.5pt, fontupper=\small] \texttt{Questions}:

\textbf{Step 1}: Clarifying the Time Frame

\textbf{Q1}: How should we interpret "last month" in the query?

\textit{Option A}: Use the previous calendar month.

\textit{Option B}: Use the last 30 days relative to today.

\vspace{0.2cm}
\textbf{Step 2}: Defining "High Spending Habits"

\textbf{Q2}: How should we determine if a customer has high spending habits?

\textit{Option A}: Consider customers whose total spending exceeds a specific monetary threshold.

\textit{Option B}: Use an existing flag or field (e.g., a "VIP" status) in the customer database.

\vspace{0.2cm}
\textbf{Step 3}: If \textit{Option A} is selected for \textit{Q2}, then:

\textbf{Q3}: Would you like to specify the spending threshold?

\textit{Option A1}: Yes, let me input a threshold value.

\textit{Option A2}: No, use a default threshold (e.g., \$1,000). \end{tcolorbox}\vspace{0.2cm}

Suppose the user selects \textit{Option A} for \textit{Q1} (previous calendar month) and \textit{Option A} for \textit{Q2} with a threshold of \$1,000 (\textit{Option A1}). The final SQL query might then be:

\begin{lstlisting}[
    language=SQL,
    basicstyle=\small\ttfamily,
    numbers=left,
    numberstyle=\tiny,
    commentstyle=\color{gray},
    breaklines=true,
    breakatwhitespace=true,
    backgroundcolor=\color{lightgrey}
]
SELECT o.order_id, o.order_date, c.customer_id, c.total_spent
FROM orders o JOIN customers c ON o.customer_id = c.customer_id
WHERE o.order_date BETWEEN '2025-02-01' AND '2025-02-28'
  AND c.total_spent > 1000;
\end{lstlisting}

The following box shows representative examples of the query’s behavior. 

\begin{tcolorbox}[breakable=true,boxsep=0pt,left=2mm,right=2mm,top=2mm,bottom=2mm, sharp corners, colback=lightblue, colframe=black, boxrule=0.5pt, fontupper=\small] 
\texttt{Example Outputs}:
\vspace{0.1cm}

\textbf{Selected:} Customer with \texttt{customer\_id} \texttt{101} has total spending \texttt{\$1,200}, and an order on 2025-02-15.
\vspace{0.2cm}

\textbf{Not Selected:} Customer with \texttt{customer\_id} \texttt{102} has total spending \texttt{\$950}, even if the order is within \texttt{February 2025}. 
\vspace{0.2cm}

\textbf{Not Selected:} Customer with \texttt{customer\_id} \texttt{103} has total spending \texttt{\$1,500}, but the order was placed on \texttt{2024-12-28}. \end{tcolorbox}\vspace{0.2cm}

This example demonstrates how the initial ambiguous prompt is transformed into a precise SQL query, and how the final query behaves with different customer data scenarios.

As a second example, we consider a dataset containing daily temperature readings. The task is to write a Python function that calculates the average temperature while excluding outlier values. The dataset is provided as a list of numeric temperature readings, and outliers should be identified using a statistical method.

Below is an example of a user-provided prompt that will be examined for ambiguities when converting natural language into a Python function. This prompt serves as the starting point of our process.

\begin{tcolorbox}[breakable=true,boxsep=0pt,left=2mm,right=2mm,top=2mm,bottom=2mm, sharp corners, colback=lightyellow, colframe=black, boxrule=0.5pt, fontupper=\small] \texttt{Prompt}: Write a Python function that calculates the average temperature from a list of readings, excluding outliers. \end{tcolorbox}\vspace{0.2cm}

The following box lists the ambiguities detected in the prompt. These ambiguities indicate areas that require further clarification.

\begin{tcolorbox}[breakable=true,boxsep=0pt,left=2mm,right=2mm,top=2mm,bottom=2mm, sharp corners, colback=lightgreen, colframe=black, boxrule=0.5pt, fontupper=\small] \texttt{Ambiguities}:

\textbf{A1}: \textit{Outlier Definition}: What criteria should be used to determine outliers (e.g., using the IQR method, z-score, or a fixed threshold)?

\textbf{A2}: \textit{Statistical Basis}: Should outliers be determined relative to the median or the mean of the dataset?

\textbf{A3}: \textit{Empty Result Handling}: What should the function return if all readings are identified as outliers? \end{tcolorbox}\vspace{0.2cm}

Next, we present a sequence of clarifying questions designed to resolve these ambiguities. Each step in the dialogue helps narrow down the possible interpretations.

\begin{tcolorbox}[breakable=true,boxsep=0pt,left=2mm,right=2mm,top=2mm,bottom=2mm, sharp corners, colback=lightpurple, colframe=black, boxrule=0.5pt, fontupper=\small] \texttt{Questions}:

\textbf{Step 1: Defining Outliers}

\textbf{Q1}: Which statistical method should be used to identify outliers?

\textit{Option A}: Use the IQR method.

\textit{Option B}: Use the z-score method.
\vspace{0.2cm}

\textbf{Step 2: Determining Statistical Basis}

\textbf{Q2}: Should outliers be determined relative to the median or the mean?

\textit{Option A}: Use the median.

\textit{Option B}: Use the mean.
\vspace{0.2cm}

\textbf{Step 3: Handling Empty Results}

\textbf{Q3}: What should the function return if all readings are excluded as outliers?

\textit{Option A}: Return \texttt{None}.

\textit{Option B}: Return 0. \end{tcolorbox}\vspace{0.2cm}

Suppose the user selects \textit{Option A} for \textit{Q1} (IQR method), \textit{Option A} for \textit{Q2} (use the median), and \textit{Option A} for \textit{Q3} (return \texttt{None}). The final Python function might then be:

\begin{lstlisting}
import numpy as np
def average_temperature(readings):
    if not readings:
        return None
    # Calculate the first and third quartiles based on the median
    q1 = np.percentile(readings, 25)
    q3 = np.percentile(readings, 75)
    iqr = q3 - q1
    lower_bound = q1 - 1.5 * iqr
    upper_bound = q3 + 1.5 * iqr
    # Filter out outliers
    filtered = [temp for temp in readings if lower_bound <= temp <= upper_bound]
    if not filtered:
        return None
    return sum(filtered) / len(filtered)
\end{lstlisting}

The following box shows representative examples of the function's behavior:

\begin{tcolorbox}[breakable=true,boxsep=0pt,left=2mm,right=2mm,top=2mm,bottom=2mm, sharp corners, colback=lightblue, colframe=black, boxrule=0.5pt, fontupper=\small] \texttt{Example Outputs}:
\vspace{0.2cm}

\textbf{Selected:} For readings \texttt{[32, 35, 36, 38, 120]}, assuming \texttt{120} is an outlier, the function returns an average of \texttt{35.25}.
\vspace{0.2cm}

\textbf{Not Selected:} For readings \texttt{[15, 16, 15, 1000]}, if \texttt{1000} is identified as an outlier, the average is computed from \texttt{[15, 16, 15]}.
\vspace{0.2cm}

\textbf{Empty Result:} For readings \texttt{[1000, 1020, 980]}, if all values are considered \texttt{outliers}, the function returns \texttt{None}. \end{tcolorbox}\vspace{0.2cm}

\subsection{Comprehensive Evaluation}
\label{sec:eval}

We conducted a comprehensive evaluation of our iterative ambiguity resolution approach using a testing set that covers a range of ambiguous prompts from different domains. All input data, including the original prompts with ambiguities, the identified ambiguities, and the corresponding disambiguated prompts, are available at \url{https://github.com/SCAlabUnical/PromptAmbiguityDataset/}. This diverse testing set was designed to assess the system’s performance in transforming vague, natural language instructions into precise and executable prompts.

Our evaluation is organized around three case studies:

\begin{enumerate} \item \textbf{Coding:}
The system transforms ambiguous natural language requests into precise code (e.g., SQL, Python) by iteratively clarifying uncertainties regarding parameters, algorithms, and data handling. This results in executable code that faithfully meets the intended requirements.

\item \textbf{Data Analysis:}
Ambiguous instructions for data analysis are refined into clear, well-defined tasks through a guided dialogue. This process resolves uncertainties about statistical measures and selection criteria, yielding robust, executable analytical scripts.

\item \textbf{Creative Writing:}
The system refines generic writing prompts—lacking details on characters, setting, or tone—into detailed instructions. Through iterative clarification, it enables the generation of coherent, creative narratives that align with the user's intent. \end{enumerate}

Our evaluation compares our iterative ambiguity resolution process with a standard one-shot output generation system. In the standard approach, users typically receive a single output from their initial prompt—often a piece of code or analysis—and then try the result; if it is not as expected, they must manually revise and resubmit new prompts to correct errors or address misunderstandings. This cycle of testing and prompt adjustment continues until the desired outcome is achieved. By contrast, our iterative process guides the user through targeted clarifications and refinements in a structured dialogue, streamlining the path to an accurate final output. As a chatbot system, we have chosen to use GPT-4o (accessed via our API) for generating responses in our iterative process and the standard one-shot version through a chat interface, although our approach is designed to be easily adaptable to other similar systems.

To thoroughly evaluate the performance and user impact of our iterative ambiguity resolution process, we focus on three key aspects:

\begin{enumerate} 
\item \textbf{Ambiguity Identification Accuracy:} We measure how effectively the system detects ambiguous elements within user prompts by calculating precision and recall against expert annotations. A higher F1-score indicates that our system reliably identifies potential ambiguities.

\item \textbf{Time Savings and Overall Productivity:} We compare the total user interaction time—from initial prompt submission to final output acceptance—as well as the number of manual corrections required by our guided approach versus the standard one-shot method. A reduction in these metrics indicates improved efficiency and productivity.

\item \textbf{Interactive Ambiguity Resolution Efficiency:} We assess user satisfaction with the guided ambiguity resolution process through questionnaires that evaluate the clarity of the generated output, the efficiency of the dialogue, and overall satisfaction with the iterative refinements. This provides insights into the benefits of structured disambiguation over conventional methods.
\end{enumerate}

For our evaluation, we engaged ten human evaluators with expertise in coding, mathematical reasoning, and statistical analysis to ensure accurate interpretation and assessment of ambiguities. The dataset consists of 75 ambiguous prompts, evenly distributed across the three case studies (25 for coding, 25 for data analysis, and 25 for creative writing). The number of ambiguities per prompt is limited and varies between 1 and 5. Each evaluator was assigned ten queries, selected randomly across different case studies, ensuring diversity in prompt evaluation. Their domain knowledge enables a reliable assessment of both the system’s ability to detect ambiguities and the effectiveness of its resolution, ensuring that the generated outputs are both precise and contextually relevant.

In the following subsections, we present the detailed results for each of these evaluation criteria, demonstrating the effectiveness of our approach through empirical analysis and user feedback.

\subsubsection{Ambiguity Identification Test}

Using our dataset, we evaluate the clarity of the ambiguities identified by our system by comparing its results against expert annotations. For each ambiguous prompt, reference ambiguities—defined and inserted into the dataset by human experts—serve as reference. The system detects a set of ambiguities for each prompt, and we compute the intersection between the system's findings and the reference. Precision is calculated as the ratio of the intersection (i.e., correctly identified ambiguities) to the total ambiguities detected by the system, while recall is the ratio of the intersection to the total ambiguities present in the reference. The F1 score is then derived as the harmonic mean of precision and recall. Note that any additional ambiguities detected by the system that are not present in the reference are not included in these calculations, even though they may be interesting and timely. Table~\ref{tab:ambiguity-results} summarizes the performance across three use cases: coding, data analysis, and creative writing.

In our evaluation, the system achieves high performance in the coding domain with a precision of 0.84, recall of 0.87, and F1 score of 0.85, reflecting the clear and structured nature of programming languages. In data analysis, the system shows a precision of 0.77, recall of 0.87, and F1 score of 0.82, indicating that it generally captures expert-identified ambiguities, though sometimes it flags additional ones. In creative writing, performance is lower (precision 0.74, recall 0.64, F1 0.69), highlighting the challenge of detecting subtle ambiguities in free-text. Overall, these results demonstrate the system's strength in structured contexts and the need for further refinement for free-text writing.

\begin{table*}[htb!]
\centering
\fontsize{8pt}{10pt}\normalfont{
\begin{tabular}{lccc} \hline \textbf{Use Case} & \textbf{Precision} & \textbf{Recall} & \textbf{F1-Score} \\ \hline 
Coding & 0.84 & 0.87 & 0.85 \\ 
Data Analysis & 0.77 & 0.87 & 0.82 \\ 
Creative Writing & 0.74 & 0.64 & 0.69 \\ \hline 
\end{tabular}
}
\caption{Ambiguity Identification Performance} 
\label{tab:ambiguity-results}
\end{table*}

\subsubsection{Time Efficiency Test}  

We measure the time users spend resolving ambiguities with our iterative approach, including the writing of the initial prompt, the system's detection of ambiguities, and the interactive clarification process between the user and the system. This process concludes with the generation of a final, corrected result. If the output is not entirely accurate, any additional refinements beyond the guided interaction are handled independently through further interactions with ChatGPT-4o. To provide a comparative analysis, we also evaluate the total number of interactions and the total time required in a standard one-shot approach until the correct output is achieved. Our evaluation is based on 10 randomly selected tests for each use case—coding, data analysis, and creative writing—to ensure a representative assessment. The aggregated averages of these results are presented in Table~\ref{tab:time-efficiency}.

\begin{table}[ht]
\centering
\fontsize{8pt}{10pt}\normalfont{
\begin{tabular}{lccccc}
\toprule
\textbf{Task Type} & 
\multicolumn{2}{c}{\textbf{Avg. number of Interactions}} & 
\multicolumn{2}{c}{\textbf{Avg. time (minutes}} \\
\cmidrule(lr){2-3} \cmidrule(lr){4-5}
 & \textbf{Standard} & \textbf{Iterative} & \textbf{Standard} & \textbf{Iterative} \\
\midrule
Coding & 6.1 & 3.0 & 17.9 & 9.0 \\
Data Analysis & 5.4 & 3.5 & 18.3 & 10.9 \\
Creative Writing & 7.2 & 5.8 & 13.8 & 10.3 \\
\bottomrule
\end{tabular}
}
\caption{Comparison of interactions and average user time between a standard one-shot approach and our iterative ambiguity resolution approach for three case studies.}
\label{tab:time-efficiency}
\end{table}

The results show that our iterative approach significantly reduces the number of interactions across all tasks, with coding and data analysis requiring approximately half as many interactions as the standard approach. This reduction is also reflected in the average time spent, where users complete tasks faster using the iterative method. In coding and data analysis, time savings are particularly notable, with reductions of approximately 50\% and 40\%, respectively. In creative writing, while the iterative approach still leads to a lower number of interactions and time spent, the reduction is less pronounced, likely due to the subjective nature of writing tasks, where refinement often requires more iteration. These findings highlight the efficiency of our approach in minimizing user effort while improving task completion time.

\subsubsection{Interactive Resolution Test}  
For each ambiguous query, we measure how many clarification iterations are needed. In one test, our system might require from one to five rounds of dialogue to resolve ambiguities, resulting in a final, correct query. In contrast, a standard system might output an incorrect query that forces the user to manually debug and revise the query over multiple attempts. User feedback is collected via questionnaires to assess clarity and satisfaction with the iterative process.

To better understand the user experience, the following questions were posed:

\begin{enumerate}
    \item How clear was the final query generated by the system?
    \item How efficient was the dialogue process in resolving ambiguities?
    \item How satisfied are you with the overall iterative process?
    \item How likely are you to recommend this system to others based on the clarification process?
    \item To what extent did the iterative clarifications improve the accuracy of the final query?
\end{enumerate}

Each question was rated on a scale of 1 to 5, with 5 being the most favorable. 

\begin{table}[ht]
\centering
\fontsize{8pt}{10pt}\normalfont{
\begin{tabular}{lc}
\toprule
\textbf{Satisfaction Question} & \textbf{Avg. Rating (out of 5)} \\
\midrule
Clarity of the Final Query            & 4.4 \\
Efficiency of the Clarification Process  & 4.8 \\
Overall Satisfaction with the Dialogue & 4.3 \\
Likelihood to Recommend the System    & 4.6 \\
Perceived Improvement from Iteration  & 4.1 \\
\bottomrule
\end{tabular}
}
\caption{User satisfaction ratings for the iterative ambiguity resolution process.}
\label{tab:questions}
\end{table}

Table~\ref{tab:questions} presents user satisfaction ratings for the iterative ambiguity resolution process. The results indicate that users found the clarification process highly efficient (4.8), suggesting that the guided approach effectively refines ambiguous prompts. The clarity of the final query (4.4) and likelihood to recommend the system (4.6) further highlight the perceived usefulness and reliability of the method. The overall satisfaction with the dialogue (4.3) reflects positive user experience, while the perceived improvement from iteration (4.1) suggests that users recognize the benefits of interactive disambiguation, though with slightly lower enthusiasm compared to other aspects. These findings demonstrate that users appreciate the structured resolution approach, reinforcing its value as an alternative to conventional one-shot methods.

\section{Conclusion}
\label{sec:conclusion}
The findings of our work demonstrate that a guided, stateful approach to prompt disambiguation significantly enhances the performance and usability of generative AI systems. By engaging users in an iterative dialogue to clarify ambiguous prompts, our method consistently yields more accurate outputs and streamlines the overall problem-solving process. This structured guidance reduces the reliance on extensive post-generation corrections, ultimately saving valuable time and effort compared to standard, free-form prompting techniques. Looking ahead, future work will focus on integrating this guided prompt methodology into a broader range of applications. We plan to explore its potential to endow generative AI systems with enhanced stateful capabilities, thereby further improving their effectiveness and user-friendliness in real-world scenarios.


\bibliographystyle{splncs04}
\bibliography{biblio}

\end{document}